\documentclass[11pt]{article}
\usepackage[preprint]{acl}
\usepackage{times}
\usepackage{latexsym}
\usepackage[T1]{fontenc}
\usepackage[utf8]{inputenc}
\usepackage{natbib}
\usepackage{microtype}

\usepackage{inconsolata}
\usepackage{graphicx}
\usepackage{latexsym}
\usepackage{amsmath}
\usepackage{stfloats}
\usepackage{multirow}
\usepackage{booktabs}
\usepackage{graphicx}
\usepackage{float}
\usepackage{hyperref}
\usepackage{url}
\usepackage{longtable}
\usepackage{tabularx}
\usepackage{adjustbox}
\usepackage{tikz}
\usetikzlibrary{shapes.geometric}
\usepackage{supertabular}

\usepackage{comment}

\title{Fine-tuning Language Models for Recipe Generation:\\A Comparative Analysis and Benchmark Study}

\author{
Anneketh Vij \thanks{Authors contributed equally.}, Changhao Liu\footnotemark[1], Rahul Anil Nair\footnotemark[1], Theodore Eugene Ho\footnotemark[1], Edward Shi, Ayan Bhowmick \\
Department of Computer Science\\
University of Southern California \\
Los Angeles, CA 90007 \\
{\tt \{anneketh, celiu, ranair, teho, epshi, abhowmic\}@usc.edu}
}

\begin{document}
\maketitle

\begin{abstract}
This research presents an exploration and study of the recipe generation task by fine-tuning various very small language models, with a focus on developing robust evaluation metrics and comparing across different language models the open-ended task of recipe generation. This study presents extensive experiments with multiple model architectures, ranging from T5-small \citep{raffel2023exploringlimitstransferlearning} and SmolLM-135M \cite{allal2024SmolLM} to Phi-2 \cite{microsoft2023phi2}, implementing both traditional NLP metrics and custom domain-specific evaluation metrics. Our novel evaluation framework incorporates recipe-specific metrics for assessing content quality and introduces approaches to allergen substitution. The results indicate that, while larger models generally perform better on standard metrics, the relationship between model size and recipe quality is more nuanced when considering domain-specific metrics. SmolLM-360M and SmolLM-1.7B demonstrate comparable performance despite their size difference before and after fine-tuning, while fine-tuning Phi-2 shows notable limitations in recipe generation despite its larger parameter count. The comprehensive evaluation framework and allergen substitution systems provide valuable insights for future work in recipe generation and broader NLG tasks that require domain expertise and safety considerations.
\end{abstract}

\section{Introduction}

The generation of safe and high-quality recipes presents unique challenges in natural language generation. Beyond generating coherent and creative recipes, recipe generation requires high-level knowledge of culinary techniques, nutritional principles, and awareness of dietary restrictions to ensure user safety. This necessitates approaches that balance linguistic fluency with domain-specific expertise, particularly in the domain of allergen substitution.

This study focuses on addressing these challenges by experimenting with different model architectures for recipe generation and allergen substitution through controlled fine-tuning and comprehensive evaluation metrics. There are three primary research questions:
\begin{enumerate}
    \item Given the scope of this study, which models will achieve the best results after fine-tuning for recipe generation?
    \item How should the generated recipes be evaluated to ensure that they are coherent and safe for users with dietary restrictions?
    \item How should allergen substitution be implemented into large-scale models to achieve high-quality allergen-free recipes?
\end{enumerate}
To answer these questions, this study makes the following contributions:
\begin{itemize}
    \item Comprehensive comparison of model architectures across scales including smaller models like GPT-2 \cite{Radford2019LanguageMA} and T5 \cite{2020t5}, and larger models like Phi-2 \cite{microsoft2023phi2} and SmolLM-1.7\cite{allal2024SmolLM}
    \item Multi-dimensional evaluation framework combining novel recipe-specific evaluation metrics, traditional metrics, and LLM-based assessment.
    \item Development of RAG-assisted and prompt-based approaches for allergen substitution
\end{itemize}
Our work represents a step forward in adapting NLG systems for practical applications in the culinary domain, emphasizing safety, personalization, and quality.
\section{Related Work}
\subsection{Language Models in Recipe Generation}
Previous works have explored various language model architectures for recipe generation. \citet{lam2024enhancing} explored the performance of BART-based \citep{lewis-etal-2020-bart} and BRIO-based \citep{liu-etal-2022-brio} across different recipe datasets, mostly in English and Vietnamese, which were then evaluated using ROUGE scores. Our work extends these previous ones by systematically comparing models across different sizes and architectures, from smaller models like SmolLM-135M to larger ones like Phi-2. This paper provides a detailed analysis on how model size impacts different aspects of recipe quality. 

\subsection{Recipe Generation Models}
This paper builds on several recent advances in recipe generation and personalization. \citet{majumder2019generatingpersonalizedrecipeshistorical} proposed a personalized recipe generation model using attention mechanisms to focus on recipes previously consumed by the user. Their approach showed promising results in generating recipes that aligned with user preferences. The dataset from this paper has been used in this research with their encoder-decoder model being used as a baseline.

\citet{Chen_2021} implemented a framework using constrained question answering over a large-scale knowledge graph to recommend food recipes considering users' explicit requirements and health factors. This helped recommend healthy alternatives to users, which aligned with the study's goal of providing allergen-free options and gave us the inspiration for a RAG-based system for allergen substitution. 

\subsection{Multi-modal Approaches}
The FIRE system, by \citet{chhikara2024firefoodimagerecipe} and Nutrify AI by \citet{han2024nutrifyaiaipoweredrealtimefood} both use a multi-modal approach, generating recipes from food images and ingredients. While it differs from this work due to us not using images, these studies also incorporate different types of input in the process of recipe generation.

LLava-Chef, by \citet{Mohbat_2024} and  is another multi-modal approach to recipe generation, which was fine-tuned on both the cross-entropy loss and a novel loss function computed using BLEU and ROUGE scores to ensure that the model generated recipes that were closer to the ground truth. This paper adopted these evaluation metrics and the idea of creating custom ones from this paper, as well as what inputs to include for recipe generation. However, this work doesn't use the novel loss function to fine-tune the language models, since penalizing generations for not being closer to the ground truth might hinder the personalization of the generated recipes, which is an important part of allergen substitution.

Other multi-modal recipe generation approaches include ChefFusion by \citet{li2024cheffusion} and Inverse Cooking by \citet{salvador2019inverse}. ChefFusion provides complete multimodality by developing a framework for both recipe generation using images and image generation using recipes. This paper, along with LLava-Chef, uses metrics like SacreBLEU\cite{post-2018-call} and ROUGE, which wasn't preferred due to the limitations of these metrics for creative generation. Inverse Cooking ,  which uses encoder-decoder transformer \cite{vaswani2017attention} blocks to generate recipes from images, provided the inspiration to use Ingredient Coverage as an evaluation metric, which is similar to how the paper evaluates the ingredients extracted from the image and the ground truth.
\vspace{-5pt}
\subsection{Evaluation}
Many of these studies, such as LLava-Chef or Retrieval Augmented Recipe Generation by \citep{liu2024retrieval}, use conventional metrics such as BLEU, ROUGE, and F1-score for ingredient matching to assess recipe quality. This paper distinguishes itself by employing both general and domain-specific metrics such as ingredient coverage, which was used by both \citet{liu2022plugandplayrecipegenerationcontent} and \citet{salvador2019inverse} to attain a more profound understanding of the quality implications across many aspects of the generated recipe since traditional metrics focus more on overlap and thus hinder creativity in generation. 

\section{Approach}
\subsection{Food.com Dataset}

The Food.com dataset \citep{majumder2019generatingpersonalizedrecipeshistorical} contains more than 180,000 recipes and 700,000 recipe reviews across 18 years. Each entry includes the recipe name, the list of ingredients, the cooking instructions, nutritional information, and user ratings and reviews. This study used the \texttt{RAW\_recipes} dataset from the Food.com dataset for research. The data preprocessing pipeline consisted of the following steps:
\begin{itemize}
    \item Extraction of recipe names, ingredients lists, and cooking instructions
    \item Standardization of ingredient formats and measurements
    \item Tokenization and formatting of recipe names, standardized ingredients and instructions
    \item Creation of input-output pairs for model training
\end{itemize}
 The format of the input is as follows:
\begin{verbatim}
<|startoftext|>[Recipe Name]
Ingredients: [Ingredients List]
\end{verbatim}
The cooking instructions were used as the target output for the models.
\subsection{Exploratory Data Analysis}
A statistical analysis of the entire dataset was conducted to gain insights into the distribution of ingredients and recipe length. 

The distribution of ingredient occurrences is dominated by a few common ingredients such as salt, butter, sugar, etc. When considering the set of unique ingredients, 9.66\% were included in 90\% of the recipes, while the remaining 91.44\% were only included in 10\% of the recipes. 

The tokenized length of recipes was also measured. 99. 4\% of the recipes had a tokenized length of less than 512 tokens and 90.4\% had less than 256. These statistics were used to determine the size of the context when training the models. 
Additional analysis can be found in Appendix \ref{sec:data_analysis}.

\subsection{Fine-Tuning Small Scale Models}

From our dataset, we randomly sampled 100,000 recipes. This was then split into training (80\%), validation (10\%), and test (10\%) sets. When evaluating the generated recipes, the first 500 samples from the test set are used to ensure consistency across different model evaluations.
We initially implemented a custom encoder-decoder model with attention, inspired by the architecture described in \citet{bahdanau2016neuralmachinetranslationjointly}. The model consisted of an embedding layer, a bidirectional GRU encoder, a GRU decoder with attention mechanism, and a final linear layer for output generation. However, this model produced near-zero scores on our evaluation metrics, indicating significant challenges in learning the complex patterns required for recipe generation.
Following the challenges with the custom model, we turned to pre-trained language models, such as SmolLM \citep{allal2024SmolLM} (135 M), GPT-2(small and medium variants)\citep{Radford2019LanguageMA}, and encoder-decoder language models like T5-small \citep{raffel2023exploringlimitstransferlearning} to explore the impact of model size and architecture on recipe generation. These models were fine-tuned on the recipe dataset, using the following approach:
\begin{itemize}
    \item Input: Combined recipe name and ingredients
    \item Output: Cooking instructions
\end{itemize}
Training configurations for the small-scale models are listed in Appendix \ref{sec: small_training-configs} and a sample output for these small-scale models is given in Appendix \ref{sec:output_small_models}.

\subsection{Fine-Tuning Larger Models}
From the evaluation metrics of the small-scale model generations, as seen in Table \ref{tab:model_comparison_small_scale}, we decided to scale up the size of our dataset to now include the entire dataset and turned towards large-scale models such as SmolLM-360M\cite{allal2024SmolLM}, SmolLM-1.7B and Phi-2 instead. We achieved this with our limited computational resources by using the QLORA approach and setting the rank to 8. The entire data set, consisting of 231637 recipes, was split into training (80\%), validation (10\%), and test (10\%) sets. As before, the first 500 samples of the test set were used for evaluation to ensure consistency between the evaluation results for the different model generations. Also, generation evaluation was now performed for both baseline and fine-tuned versions to better understand the impact fine-tuning had on the generated recipes. The training configurations of these large-scale models are listed in Appendix \ref{sec:large-training-configs}, and all models were trained on 1 epoch on these configurations for 8 hours on 2 NVIDIA A100 GPUs.

\subsection{Allergen Substitution}
Allergen substitution was performed when generating recipes using the following two approaches:
\subsubsection{Prompt based Allergen Substitution}
Since we had fine-tuned three large-scale models on the entire data set, we hypothesized that these models should be powerful enough to substitute the allergens present in the generated recipe just by prompting the model. This was done by adding a list of allergens to avoid in the prompt along with the recipe name and the ingredient list. In order to test this approach, some common allergens, such as milk, eggs, and fish, are added to a list of allergens to avoid in the prompt. The prompt is given as follows:
\begin{quote}
   ``You are an expert chef and recipe writer with a deep understanding of culinary techniques and food allergies. Your goal is to create a detailed and high quality recipe that uses the provided list of ingredients, while making substitutions for any allergens to ensure the recipe is safe for individuals with those allergies.
Please follow these instructions:
\begin{enumerate}
\item {Create a Recipe}: Write a full, detailed recipe based on the name and ingredients provided.
\item {Substitute Allergens}: Some people are allergic to certain ingredients. You must avoid these allergens in the recipe and suggest substitutions from the list of safe ingredients. If the allergen is an essential part of the recipe, ensure the substitute maintains the flavor and texture as much as possible.
\item {Ensure Clarity and Detail}: Provide precise instructions, including cooking methods, preparation steps, and any necessary tips. The recipe should be easy to follow for someone with basic cooking knowledge.
\end{enumerate}
Create a recipe for: {name}\\
Using these ingredients: {ingredients}\\
Substitute these allergens for other ingredients: {allergens}\\
Recipe:''

\end{quote}
A sample output for these models with and without allergen substitution is given in Appendix \ref{sec:sample-recipes}. The hyperparameters for the prompt-based model is given in Appendix \ref{sec:prompt-hyperparameters}.
\subsubsection{RAG-assisted Allergen Substitution System}
The second approach was to develop an experimental RAG-assisted allergen substitution system \cite{lewis2021retrievalaugmentedgenerationknowledgeintensivenlp} to replace allergens in the generated recipes with similar ingredients as mentioned in an allergen database that we built.
Key components include:
\begin{itemize}
    \item FAISS vector store for efficient similarity search
    \item HuggingFace embeddings (sentence-transformers/all-MiniLM-L6-v2)
    \item Custom allergen database with substitution rules
    \item Ingredient parsing and validation system
\end{itemize}
Implementation details:
\begin{itemize}
    \item Chunk size: 1000 tokens
    \item Chunk overlap: 200 tokens
    \item Top-k retrieval: k=1 for substitution matches
\end{itemize}
A workflow for the RAG-assisted system can be found in Appendix \ref{sec:rag-implementation}, and the hyperparamters can be found in Appendix \ref{sec:rag-hyperparameters}. The system finds the ingredients present in the generated recipe and, if they are present in the allergen database, substitutes them with an appropriate ingredient from the database. This allergen ingredient database can be seen in Appendix \ref{sec: allergen-database}. A sample output for these models with and without allergen substitution is given in Appendix \ref{sec:RAG-Generation}.
\section{Evaluation Metrics}
A comprehensive evaluation framework has been implemented to evaluate the recipes generated by these models. The metrics can be divided into three parts.
\subsection{Traditional NLP Metrics}
This work uses traditional NLP metrics to evaluate the quality of the generated recipes.
\begin{enumerate}
    \item BLEU (Bilingual Evaluation Understudy) \cite{papineni2002bleu} is a metric that evaluates the generated text by comparing it with the ground truth. It compares the n-grams between the generated recipe and the ground truth recipe, assigning a score between 0 and 1.
    \item ROUGE (Recall-Oriented Understudy for Gisting Evaluation) \cite{lin2004rouge}, is a metric that evaluates the generated recipe by comparing the overlap between the generated recipe and the ground truth. This study will use ROUGE 1, ROUGE 2 and ROUGE L for evaluation.
    \item Perplexity is another traditional NLP metric that is used to measure the quality of the generated text. It is calculated as the exponentiated average negative log-likelihood of a sequence.
\end{enumerate}
\subsection{Recipe Specific Auto Evaluation Metrics}
The traditional metrics above are good for measuring overlap with the ground truth. However, they do not work well for evaluating a creative task such as generating recipes. A high quality generated recipe could be given a low score because it does not have much overlap with the ground truth. Therefore, we have implemented custom auto-evaluation metrics which are tailored to evaluate the generated recipes in various subdomains.
\begin{enumerate}
    \item Ingredient Coverage Tracking: Measures how effectively the generated recipe utilizes the input ingredients. It tokenizes the ingredient list, matches the ingredients in the generated instructions, and then calculates the coverage ratio, which is the number of present ingredients divided by the total number of ingredients. The metric can handle several variations and forms.
    \item Step Complexity: Evaluates instruction completeness and detail. This is done by counting the distinct operations, analyzing the step length and detail, evaluating the parameter specifications, and then calculating the complexity score.
    \item Recipe Coherence: Assesses the logical flow and structure of the recipe. This is done by building a step dependency graph, verifying the logical ordering, checking the temporal consistency, and finally calculating the coherence score.
    \item Temperature/Time Specification Checks:- Verifies critical cooking parameters by extracting the numerical values of temperature and time in the generated recipe, validating the ranges per method, checking the completeness, and then calculating the final score.
\end{enumerate}
All of these metrics result in scores between 0 and 1, where the higher the score, the better. A more detailed explanation of these metrics can be found in Appendix \ref{sec: domain-metrics}.
\subsection{LLM-As-A-Judge}
This work also uses the LLM-as-a-judge method to evaluate the recipes generated by the baseline and fine-tuned versions of the models. Initially, we used Qwen2.5-1.5B Instruct \cite{qwen2} \cite{qwen2.5}, but shifted to a much larger model in Qwen2.5-7B \cite{qwen2.5} for more accurate scores when judging the quality of the generated recipes. The recipes are evaluated using six Likert scale categories and are scored on a scale of 1-5. These categories are as follows:
\begin{enumerate}
    \item Clarity: Instruction comprehensibility
    \item Completeness: Coverage of necessary steps
    \item Consistency: Logical flow and coherence
    \item Practicality: Feasibility of execution
    \item Relevance: Alignment with recipe goals
    \item Allergen Safety: Checks if allergen is substituted correctly
\end{enumerate}
\section{Results}
\subsection{Initial Results with Small-Scale Models}
Table \ref{tab:model_comparison_small_scale} presents the initial results, comparing the recipes generated with small-scale models, using BLEU and ROUGE metrics. 
\begin{table}[h]
\centering
\begin{adjustbox}{width=\columnwidth,center}
\begin{tabular}{lrrrrrr}
\toprule
Model & ROUGE-1 & ROUGE-2 & ROUGE-L & BLEU-1 & BLEU-2 & BLEU-3 \\
\midrule
Custom Encoder-Decoder & 0.10 & 0.02 & 0.08 & 0.05 & 0.01 & 0.00 \\
SmolLM (Fine-tuned)    & 0.22 & 0.03 & 0.11 & 0.15 & 0.04 & 0.01 \\
GPT-2 (Small)          & 0.25 & 0.05 & 0.15 & 0.18 & 0.07 & 0.03 \\
GPT-2 Med              & 0.28 & 0.06 & 0.17 & 0.20 & 0.08 & 0.04 \\
GPT-2 Med (Fine-Tuned) & 0.33 & 0.07 & 0.19 & 0.25 & 0.11 & 0.06 \\
T5-Small (Fine-tuned)  & 0.13 & 0.04 & 0.11 & 0.00 & 0.00 & 0.00 \\
\bottomrule
\end{tabular}
\end{adjustbox}
\caption{Comparison of recipes generated by various small-scale models}
\label{tab:model_comparison_small_scale}
\end{table}
\subsection{Results with Large-Scale Models}
Table \ref{tab:model_comparison_traditional_metrics} and Table \ref{tab:model_comparison_domain_specific_metrics} contain the evaluation scores of the baseline and fine-tuned versions of the large-scale models for both traditional NLP metrics and domain-specific auto-evaluation metrics. As mentioned above, the models have low BLEU and ROUGE scores due to there not being much overlap with the ground truth, hence the use of the domain-specific evaluation metrics.

\renewcommand{\arraystretch}{0.7}
\begin{table*}[t!]
\centering
\begin{adjustbox}{width=\textwidth}
\begin{tabular}{lrrrrrrrr}
\toprule
Model & ROUGE-1 & ROUGE-2 & ROUGE-L & BLEU-1 & BLEU-2 & BLEU-3 & BLEU4 & Perplexity \\
\midrule
SmolLM (360M) - Baseline & 0.13 & 0.01 & 0.07 & 0.08 & 0.02 & 0.01 & 0.00 & 125.2 \\
SmolLM (360M) - Finetuned & 0.11 & 0.01 & 0.06 & 0.07 & 0.01 & 0.01 & 0.00 & \textbf{90.67} \\
\midrule
SmolLM (1.7B) - Baseline & 0.14 & 0.01 & 0.07 & 0.08 & 0.02 & 0.01 & 0.00 & 171.07 \\
SmolLM (1.7B) - Finetuned & 0.11 & 0.01 & 0.05 & 0.07 & 0.01 & 0.00 & 0.00 & \textbf{112.13} \\
\midrule
Phi-2 - Baseline  & 0.22 & 0.03 & 0.10 & 0.14 & 0.05 & 0.02 & 0.01 & 58.74 \\
Phi-2 - Finetuned & 0.17 & 0.01 & 0.07 & 0.11 & 0.03 & 0.01 & 0.00 & 78.9 \\
\bottomrule
\end{tabular}
\end{adjustbox}
\caption{Comparison of Large Scale Models using Traditional Metrics}
\label{tab:model_comparison_traditional_metrics}
\end{table*}

\begin{table*}
\centering
\begin{adjustbox}{width=\textwidth}
\begin{tabular}{lrrrr}
\toprule
Model & Ingredient Coverage & Step Complexity & Recipe Coherence & Temp. and Time Spec.\\
\midrule
SmolLM (360M) - Baseline & 0.21 & 0.93 & 0.03 & 0.10 \\
SmolLM (360M) - Finetuned & 0.16 & \textbf{0.98} & 0.02 & \textbf{0.12} \\
\midrule
SmolLM (1.7B) - Baseline & 0.29 & 0.84 & 0.05 & 0.11 \\
SmolLM (1.7B) - Finetuned & 0.27 & \textbf{0.97} & 0.04 & 0.03 \\
\midrule
Phi-2 - Baseline  & 0.59 & 0.79 & 0.08 & 0.329 \\
Phi-2 - Finetuned & 0.30 & \textbf{0.99} & 0.07 & 0.24 \\
\bottomrule
\end{tabular}
\end{adjustbox}
\caption{Comparison of Large Scale Models Using Domain Specific Metrics}
\label{tab:model_comparison_domain_specific_metrics}
\end{table*}
\begin{table*}[!t]
\centering
\begin{adjustbox}{width=\textwidth}
\begin{tabular}{lrrrr}
\toprule
Model & Ingredient Coverage & Step Complexity & Recipe Coherence & Temp. and Time Spec.\\
\midrule
SmolLM (360M) - Baseline & 0.13 & 0.74 & 0.04 & 0.13\\
SmolLM (360M) - Finetuned & 0.11 & \textbf{0.92} & 0.03 & 0.09\\
\midrule
SmolLM (1.7B) - Baseline & 0.15 & 0.77 & 0.06 & 0.13\\
SmolLM (1.7B) - Finetuned & \textbf{0.16} & \textbf{0.91} & 0.05 & 0.07\\
\midrule
Phi-2 - Baseline  & 0.30 & 0.82 & 0.09 & 0.20\\
Phi-2 - Finetuned & 0.18 & \textbf{0.99} & 0.08 & \textbf{0.21}\\
\bottomrule
\end{tabular}
\end{adjustbox}
\caption{Comparison of Prompt-based Allergen Substitution using Domain Specific Metrics}
\label{tab:model_comparison_llm_metrics}
\end{table*}
\renewcommand{\arraystretch}{1}

\subsection{Results of Prompt-based Allergy Substitution}
Table \ref{tab:model_comparison_llm_metrics} contains the domain-specific auto-evaluation metrics of the baseline and fine-tuned versions of the large-scale models using prompt-based allergy substitution.  
Table \ref{tab:model_comparison_llm} shows the results of the evaluation using Qwen2.5-7B as a judge for the allergen-substituted recipes generated by the baseline and fine-tuned versions of the models. Evaluation is performed on the first 500 samples of the test set. The radar charts of these results are given in Appendix \ref{sec: llm-prompt}.

\subsection{Results of RAG-Assisted Allergy Substitution}

Table \ref{tab:model_comparison_rag_llm_metrics} contains the domain-specific auto-evaluation metrics of the baseline and fine-tuned versions of the large-scale models with RAG-assisted allergy substitution. Table \ref{tab:model_comparison_rag_llm} contains the results of the evaluation conducted by Qwen2.5-7B as a judge. As before, evaluation for the LLM-as-a-judge is performed on the first 500 samples of the test set. The radar charts of these Qwen2.5-7B results are given in Appendix \ref{sec: llm-rag}.

\section{Discussion}
Our comprehensive evaluation across model architectures and scales reveals several profound insights about the intersection of recipe generation and allergen awareness, challenging conventional assumptions about model scaling and domain adaptation. 

\begin{enumerate}
\item\textbf{Comparison between Recipe Generation and Allergen Substitution Generation:} In the domain-specific metrics, the recipes generated by the large-scale models had higher step complexity and ingredient coverage compared to the recipes generated by the prompt-based and RAG-assisted methods. There were improvements, albeit marginal ones, in overall recipe coherence in the allergen-substituted generations versus the normal generations, signifying a comparatively smaller trade-off between step complexity and other metrics. This decrease in performance for the prompt-based method is most likely due to changes in the prompt, i.e., asking the model to substitute allergens, overwhelming the model and preventing it from generating high-quality recipes. For the RAG-assisted method,  the change in hyperparameters, where the top p-value was lowered to allow the substitution of ingredients, inadvertently resulted in lower-quality recipes.
\item\textbf{Fine-tuning Dynamics:} The most interesting findings come from the fine-tuned models. For instance, despite its sophisticated architecture, Phi-2 exhibited unexpected behavior post-fine-tuning. While the baseline model achieved high scores in ingredient coverage (0.59) and temperature specification (0.329), the fine-tuned version showed significant degradation across multiple metrics. Although the fine-tuned version showed a remarkable improvement in step complexity (0.82 to 0.99), Phi-2 noticeably showed degradation in the other three metrics, suggesting that its improvement in generating recipes in a complete step-by-step manner is done by trading off semantic relations within the instructions. A similar trend was also observed in the other models to a lesser extent. This shows that conventional fine-tuning approaches may need to be revised for larger models in specialized domains. 
\item\textbf{Allergen Substitution and Evaluation Framework:} The prompt-based substitution system revealed trade-offs between safety and culinary creativity. The fine-tuned SmolLM models, both 360M and 1.7B, demonstrated promising results in allergen safety (scores of 2.57 and 2.54), although these improvements came at the cost of recipe coherence, similar to the Phi-2 models. The multi-dimensional evaluation approach revealed significant discrepancies between traditional metrics and practical applicability, as seen by Phi-2's metrics in both prompt-based and RAG-assisted allergen substitution.
\item \textbf{Comparison between Prompt-based and RAG-assisted Allergen Substitution Systems :} For domain-specific metrics,  the RAG-assisted method had higher scores in step complexity and temperature and time specification in all three models compared to the prompt-based method, with similar scores in recipe coherence and lower scores in ingredient coverage. The lower scores are most likely due to the RAG-assisted method having more ingredients to substitute. The increased scores in step complexity and temperature and time specification are most likely due to the prompt-based method struggling to generate a step-by-step recipe when allergens are present in the recipe, whereas the RAG-assisted approach only needs to substitute allergens in the generated recipe. We also find that for the LLM-as-a-judge metric, the prompt-based method outperforms the RAG-assisted method across all models and metrics. This shows that allergen substitutions alone will not produce high-quality recipes, hence the lower scores.
\end{enumerate}

\section{Future Work}
Based on the findings in this paper, we identify several promising directions to advance recipe generation with allergen awareness. 
\begin{enumerate}
\item The performance degradation observed in larger models during fine-tuning calls for more sophisticated adaptation approaches. Future work should explore constitutional fine-tuning techniques that better preserve model capabilities while adapting to the culinary domain, complemented by specialized pre-training objectives incorporating culinary domain knowledge. We envision a multi-task learning framework that simultaneously optimizes for recipe quality and allergen safety. 
\item Future work should explore other datasets and consider using multiple datasets for fine-tuning, as well as focus on better evaluation metrics for evaluation of the generated recipes.
\item The RAG-assisted allergen substitution system shows promise, but requires further development. Future research should focus on integrating comprehensive domain-specific knowledge bases for more accurate substitutions, with real-time validation mechanisms ensuring substitution safety while maintaining recipe coherence. 
\end{enumerate}

\renewcommand{\arraystretch}{0.7}
\begin{table*}[!t]
\centering
\begin{adjustbox}{width=\textwidth}
\begin{tabular}{lrrrrrr}
\toprule
Model & Clarity & Completeness & Consistency & Practicality & Relevance & Allergen Safety\\
\midrule
SmolLM (360M) - Baseline & 2.35 & 2.4 & 2.26 & 2.47 & 3.02 & 2.26\\
SmolLM (360M) - Finetuned & \textbf{2.46} & \textbf{2.6} & 2.114 & 2.28 & 2.84 & \textbf{2.57} \\
\midrule
SmolLM (1.7B) - Baseline & 2.38 & 2.42 & 2.26 & 2.48 & 3.01 & 2.29 \\
SmolLM (1.7B) - Finetuned & \textbf{2.42} & \textbf{2.57} & 2.1 & 2.28 & 2.96 & \textbf{2.54} \\
\midrule
Phi-2 - Baseline  & 2.61 & 2.54 & 2.48 & 2.71 & 3.04 & 2.46\\
Phi-2 - Finetuned & 2.29 & 2.24 & 2.01 & 2.04 & 2.32 & 2.44\\
\bottomrule
\end{tabular}
\end{adjustbox}
\caption{Comparison of Prompt-based Allergen Substitution using Qwen2.5-7b}
\label{tab:model_comparison_llm}
\end{table*}
\begin{table*}
\centering
\begin{adjustbox}{width=\textwidth}
\begin{tabular}{lrrrr}
\toprule
Model & Ingredient Coverage & Step Complexity & Recipe Coherence & Temp. and Time Spec.\\
\midrule
SmolLM (360M) - Baseline & 0.11 & 0.91 & 0.03 & 0.12\\
SmolLM (360M) - Finetuned & 0.09 & \textbf{0.98} & 0.02 & \textbf{0.13}\\
\midrule
SmolLM (1.7B) - Baseline & 0.13 & 0.83 & 0.06 & 0.16\\
SmolLM (1.7B) - Finetuned & 0.13 & \textbf{0.97} & 0.06 & 0.04\\
\midrule
Phi-2 - Baseline  & 0.34 & 0.82 & 0.08 & 0.37\\
Phi-2 - Finetuned & 0.16 & \textbf{0.99} & \textbf{0.12} & 0.26\\
\bottomrule
\end{tabular}
\end{adjustbox}
\caption{Comparison of Rag-Assisted Allergen Substitution using Domain Specific Metrics}
\label{tab:model_comparison_rag_llm_metrics}
\end{table*}
\begin{table*}[!t]
\centering
\begin{adjustbox}{width=\textwidth}
\begin{tabular}{lrrrrrr}
\toprule
Model & Clarity & Completeness & Consistency & Practicality & Relevance & Allergen Safety\\
\midrule
SmolLM (360M) - Baseline & 2.206 & 2.188 & 2.065 & 2.16 & 2.42 & 2.172\\
SmolLM (360M) - Finetuned & 2.167 & 2.112 & 1.945 & 1.97 & 2.211 & \textbf{2.283} \\
\midrule
SmolLM (1.7B) - Baseline & 2.250 & 2.246 & 2.095 & 2.188 & 2.511 & 2.251 \\
SmolLM (1.7B) - Finetuned & \textbf{2.31} & \textbf{2.28} & \textbf{2.101} & 2.125 & 2.43 & \textbf{2.413} \\
\midrule
Phi-2 - Baseline  & 2.335 & 2.342 & 2.266 & 2.368 & 2.503 & 2.273\\
Phi-2 - Finetuned & 2.146 & 2.084 & 1.998 & 2.061 & 2.229 & 2.163\\
\bottomrule
\end{tabular}
\end{adjustbox}
\caption{Comparison of Rag-Assisted Allergen Substitution using Qwen2.5-7b}
\label{tab:model_comparison_rag_llm}
\end{table*}
\renewcommand{\arraystretch}{1}

\section{Conclusion}
This work presents a comprehensive exploration of recipe generation and allergen substitution, demonstrating both the possibilities and challenges in developing practical AI systems for culinary applications. Our systematic evaluation across multiple model scales and architectures provides valuable insights into the relationship between model capacity and domain-specific performance. Our results highlight three key findings. 
\begin{enumerate}
\item There was a comparatively lower trade-off between step complexity and other metrics observed in the recipes generated by the allergen substitution systems compared to the normal generations by the large-scale models. The lower performance of the allergen substitution systems in the domain-specific metrics can be attributed to the allergen substitution prompt demanding too much from the model and the tweaking of hyperparameters allowing lower-quality recipes to be generated.
\item The challenge of maintaining recipe quality while implementing allergen substitutions requires careful balancing, as shown by the prompt-based substitution results and validated through an LLM-based evaluation. Merely substituting the allergen for a different ingredient, as shown in the RAG-assisted method, is not sufficient to solve this problem. 
\item The multi-dimensional evaluation framework reveals that traditional NLP metrics alone are insufficient for assessing recipe generation quality, emphasizing the need for domain-specific metrics. The performance degradation, particularly in fine-tuning larger models and implementing reliable allergen substitutions, should be the main focus for future developments in recipe generation systems.
\end{enumerate}
Ultimately, this work contributes to the broader field of natural language generation by demonstrating that successful recipe generation systems must balance multiple objectives: linguistic coherence, culinary accuracy, and safety considerations. These insights extend beyond recipe generation to inform the development of other domain-specific LM's where safety and expertise are paramount.
\section{Limitations}
\begin{enumerate}
    \item \textbf{Computation resources:} This study examines comparatively smaller models trained for 1-2 epochs. Future research can extend this work by exploring larger models and training for more epochs to enhance performance and robustness.
    \item \textbf{Using LLM-As-A-Judge for Evaluation:} LLM-as-a-judge is quite stochastic and computationally expensive in terms of generating scores for each recipe. Future research should focus on improving the trustworthiness of LLM-based evaluation and the efficient calculation of scores.
    \item \textbf{Evaluation on only a part of test set:} Evaluation was only performed on 500 of the generated recipes from the test set. Future research could expand the sample size to improve statistical significance and generalizability.
    \item \textbf{Language of the dataset used:} The dataset used for this research is predominantly in English, as are the generated recipes. Future research can focus on expanding to incorporate datasets in multiple languages for recipe generation.
\end{enumerate}
\bibliography{custom}
\appendix
\renewcommand{\thesection}{\Alph{section}}

\section{Exploratory Data Analysis}
\label{sec:data_analysis}
\begin{figure}[!h]
    \centering
    \includegraphics[width=.45\textwidth]{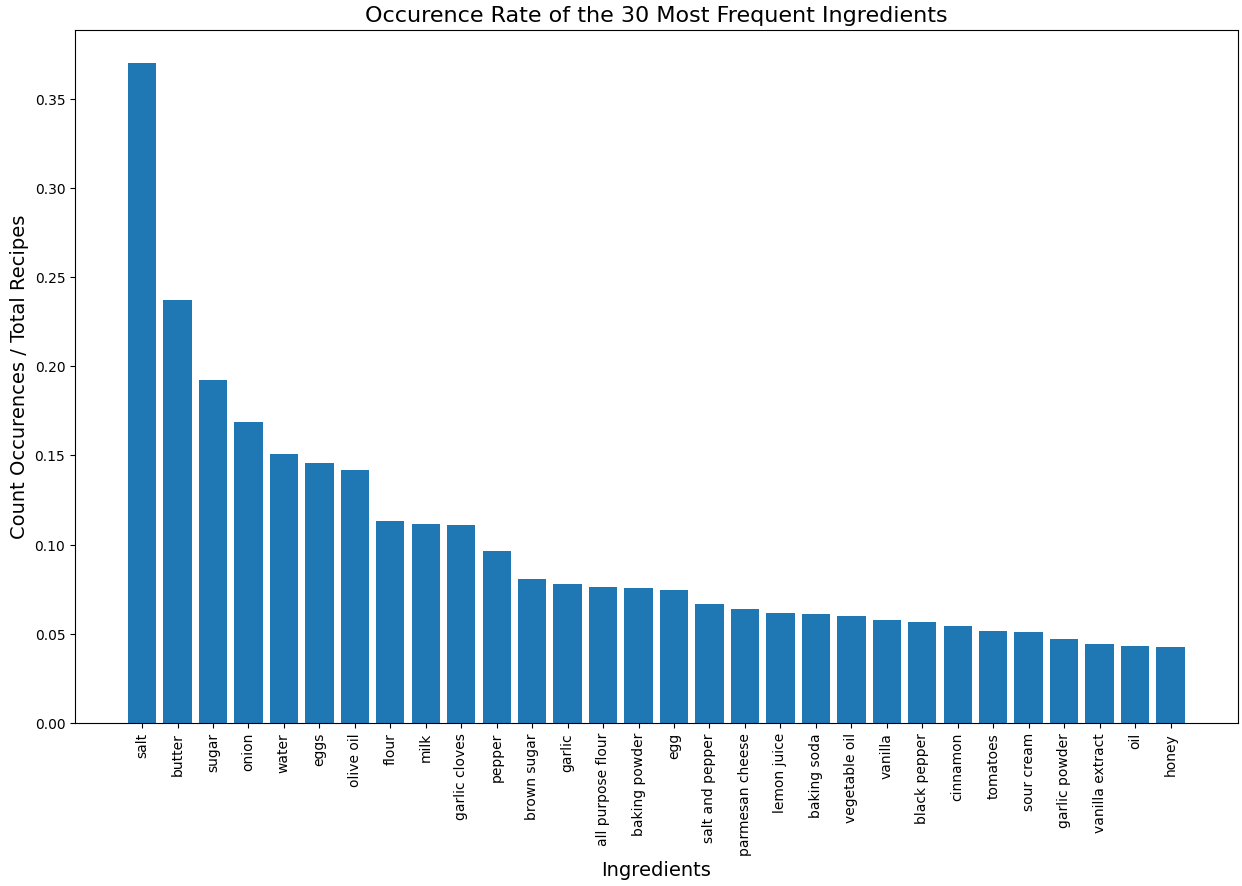}
    \caption{Occurrence rate of the 30 most Frequent Ingredients}
    \label{fig:occurrence_rate_ingredients}
\end{figure}
\begin{figure}[!h]
    \centering
    \includegraphics[width=.45\textwidth]{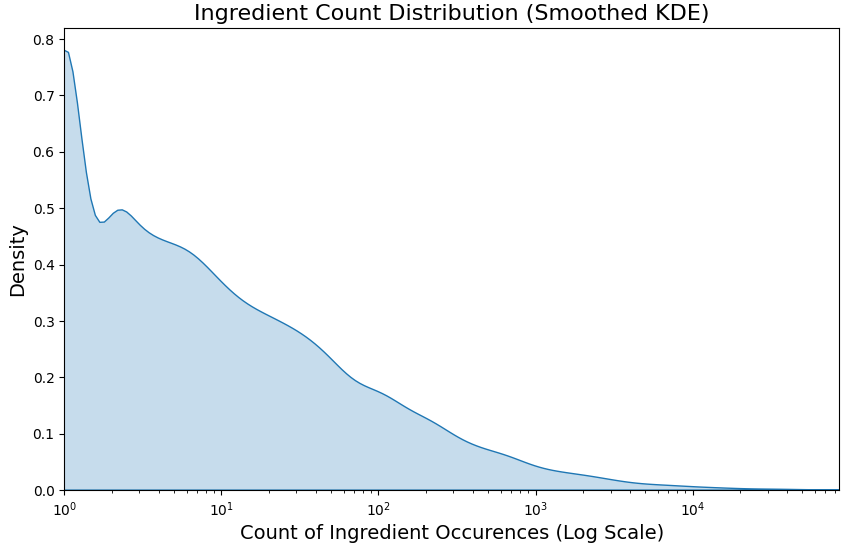}
    \caption{Ingredient Count Distribution (Smoothed KDE)}
\end{figure}
\begin{figure}[!h]
    \centering
    \includegraphics[width=.45\textwidth]{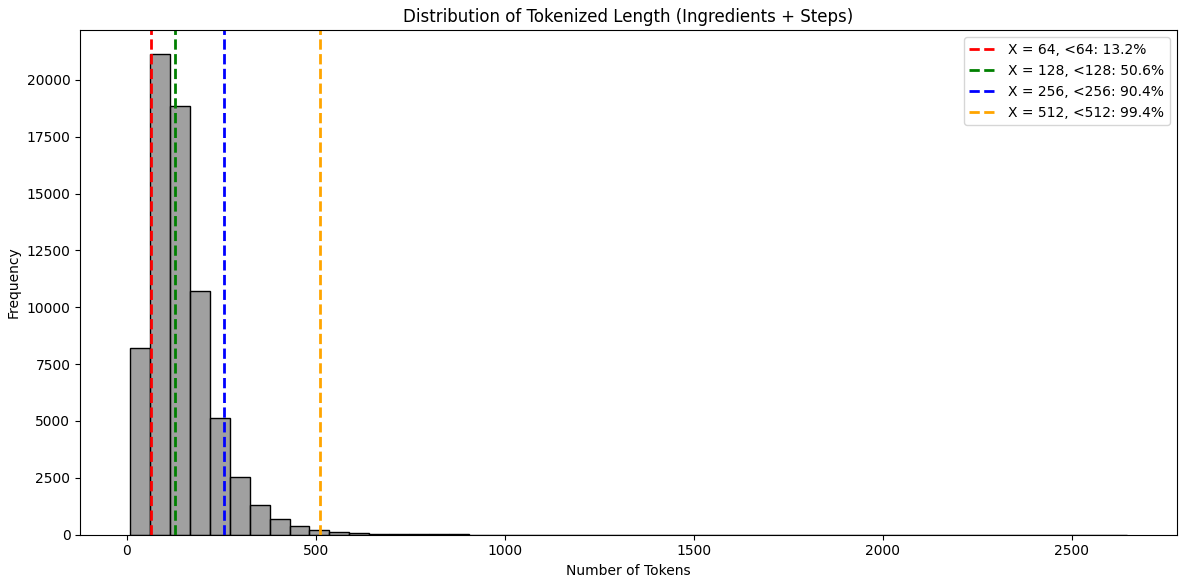}
    \caption{Distribution of Tokenized Length (Ingredient+Steps)}
    \label{fig:ingredient_count_distribution}
\end{figure}
\vspace{200pt}
\section{Sample Output for Small-Scale Model}
\label{sec:output_small_models}
\begin{table}[!h]
    \centering
    \scriptsize
    \begin{tabular}{|l|p{4.5cm}|}
        \hline
        \textbf{Section} & \textbf{Content} \\
        \hline
        Input & Chocolate Chip Cookies\newline Ingredients: flour, sugar, butter, chocolate chips, eggs, vanilla extract, baking soda, salt \\
        \hline
        Baseline Model Output & Preheat oven to 350°F. In a large bowl, cream together the butter and sugar until light and fluffy. Beat in the eggs one at a time, then stir in the vanilla. Combine the flour, baking soda, and salt; gradually stir into the creamed mixture. Fold in the chocolate chips. Drop by rounded tablespoons onto ungreased baking sheets. Bake for 8 to 10 minutes or until golden brown. Cool on wire racks. \\
        \hline
        Fine-tuned Model Output & Preheat the oven to 375°F (190°C). Line a baking sheet with parchment paper. In a large bowl, cream together the butter and sugar until light and fluffy, about 3-4 minutes. Beat in the eggs one at a time, then stir in the vanilla extract. In a separate bowl, whisk together the flour, baking soda, and salt. Gradually add the dry ingredients to the butter mixture, mixing until just combined. Fold in the chocolate chips. Drop rounded tablespoons of dough onto the prepared baking sheet, spacing them about 2 inches apart. Bake for 9-11 minutes, or until the edges are golden brown. Allow cookies to cool on the baking sheet for 5 minutes before transferring to a wire rack to cool completely. \\
        \hline
    \end{tabular}
    \caption{Comparison of Baseline and Fine-tuned Model Outputs for Chocolate Chip Cookie Recipe}
    \label{table:cookie_recipe_comparison}
\end{table}
\clearpage
\section{RAG-assisted Allergen Substitution System Workflow}
\label{sec:rag-implementation}
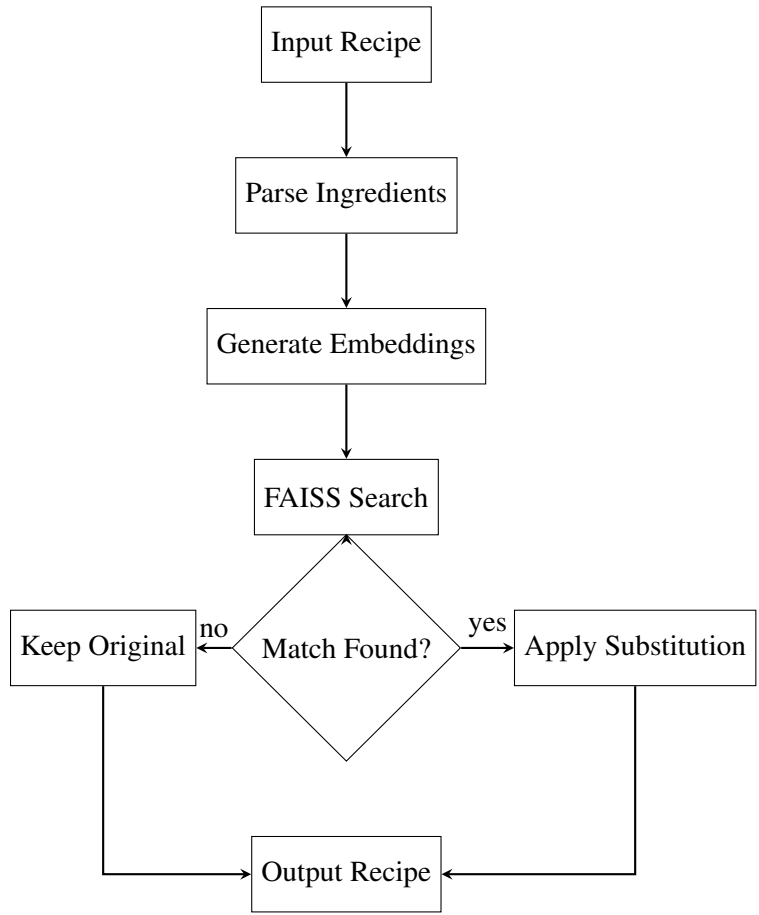
\begin{figure}[H]
\centering
\begin{tikzpicture}[node distance=2cm]
\tikzstyle{process} = [rectangle, minimum width=2cm, minimum height=1cm, text centered, draw=black]
\tikzstyle{decision} = [diamond, minimum width=2cm, minimum height=1cm, text centered, draw=black]
\tikzstyle{arrow} = [thick,->,>=stealth]

\node (start) [process] {Input Recipe};
\node (parse) [process, below of=start] {Parse Ingredients};
\node (embed) [process, below of=parse] {Generate Embeddings};
\node (search) [process, below of=embed] {FAISS Search};
\node (match) [decision, below of=search] {Match Found?};
\node (sub) [process, right of=match, xshift=1.8cm] {Apply Substitution};
\node (skip) [process, left of=match, xshift=-1.2cm] {Keep Original};
\node (end) [process, below of=match, yshift=-1cm] {Output Recipe};

\draw [arrow] (start) -- (parse);
\draw [arrow] (parse) -- (embed);
\draw [arrow] (embed) -- (search);
\draw [arrow] (search) -- (match);
\draw [arrow] (match) -- node[anchor=south] {yes} (sub);
\draw [arrow] (match) -- node[anchor=south] {no} (skip);
\draw [arrow] (sub) |- (end);
\draw [arrow] (skip) |- (end);

\end{tikzpicture}
\caption{Experimental RAG-based Allergen Substitution System Workflow}
\label{fig:rag-workflow}
\end{figure}
\section{Sample Generated Recipes from Prompt-based Method}
\label{sec:sample-recipes}
The following table shows recipe generations given the following parameters: "name": "cottage cheese waffles", "ingredients": "flour, salt, eggs, milk, vegetable oil, vanilla extract, cottage cheese", and "allergens": ["milk", "eggs", "fish"]. 
\onecolumn
\begingroup
\small
\begin{longtable}{|l|p{0.4\textwidth}|p{0.4\textwidth}|}
\hline
\textbf{Model} & \textbf{Original Output} & \textbf{With Allergen Substitution} \\
\hline
\endfirsthead
\hline
\multicolumn{3}{|c|}{{\bfseries \tablename\ \thetable{} -- continued from previous page}} \\
\hline
\textbf{Model} & \textbf{Original Output} & \textbf{With Allergen Substitution} \\
\hline
\endhead
\hline
\multicolumn{3}{|c|}{\textit{Continued on next page}} \\
\hline
\endfoot
\hline
\endlastfoot
\hline
SmolLM-360M: & 350 fahrenheit oven cake rolls or cupcakes. 2 cups all purpose whole wheat baking mix caster sugar butter egg white nonfat sour cream almond meal cocoa powder cinnamon pure maple syrup water orange juice lime zest organic dark chocolate nutmeg bourbon whiskey unsweetened condensed sweetened applesauce brown sugar coconut husk powdered sweetener granulated cane soda cornstarch finely ground cardamom peppercorns black pepper dry ice (for frosting) light whipping machine cold kosher buttermilk melted margarine whipped topping blueberry sauce frozen raspberries lemon rind green apple red onion fresh parsley mint jalapeno chili peppers freshly grated pepitas garlic cloves sea bass flakes mayo diced tomatoes chopped cilantro chile relish sweet potato hash chips ham hock bacon seasoning & Soft Serve Waffle Cups with Creamy Cottage Cheese Topping \newline Bake at 400\textdegree F (around 256\textdegree C) until golden brown around edges \& bubbly inside - about 8 minutes or so after turning oven light/heat setting off when ready . Add cream yogurt mix , coconut flakes if desired ! Slice into squares approx., then cut napkins out diagonally just before serving! Enjoy your delicious meal made healthier too :)\\ 
\hline
SmolLM-1.7B & Cottage Cheese Wafels with Vanilla Ice Cream Topping. 1 cup all purpose flour 2 teaspoons kosher baking powder pinches nutmeg pepper fresh ground black pepper cold water butter sugar egg yolk unsalted salted reduced fat creamer whole grain oats half \& halves lowfat cheddar or Swiss cheeses light brown breadcrumbs apple juice frozen whipped topping ice cubes chocolate syrup optional garnish cooked carrots celery cucumbers green onions lemon wedges whipping whip heavy cream strawberry puree sherry wine fruit cocktail soda strawberries blueberries raspberries peaches pears bananas grapes watermelon seedless grapefruits pineapples orange slices mandarin oranges pineapple chunks cherries apples tart cherry pomegranate seeds extra virgin olive spread dried cranberries cinnamon crushed nuts coconut marshmallows angel food cake frosting cocoa dusted powdered sugar graham crackers swirl candy melts sea salt dark rum soy sauce maple whiskey lime zest fennel bulb dill weed shallot scallions yellow onion garlic cloves mint leaves basil chives parsley thyme oregano saffron turmeric coriander cumin cardamom ginger marjoram bay leaf peppercorns rose petals star anise licorice root tumeric goldenrod geranium tea clove rosemary elderberry spice pumpkin pie spices jalapeno peppers red bell peppers canned mushrooms chicken breast boneless skinless turkey wing tenderlo & Cottage Cheese Waffle Bites - Instructions per person: 1) Preheat oven to 400\textdegree F (around room temperature). Grease two baking sheets lightly using nonstick spray or butter . Melt some coconut oil over medium heat , add chopped onion \& garlic if desired ; cook until softened slightly but not browned): Prepare batter by combining dry mixes [flour / oatmeals] separately then whisk together wet mixes[water + melted fat+ mixed vegetables]. Pour equal amounts into individual ramekins spread evenly apart onto greased sheet trays ensuring no overflowing occurs during boiling phase later b): Once bubbles appear around edges let rest about five minutes before adding more liquid gradually stirring well after every addition till smooth consistency achieved without lumps remaining c ): Drop spoonful sized portions gently off counter surface directly upwards through small hole made earlier within parchment lined tray avoiding excessive air exposure between drops causing deflation affecting final product shape uniformity upon serving time ! Enjoy warm immediately post completion! Good luck crafting your delicious creation today :) Happy Cooking!! \\
\hline
Phi-2 & Cottage Cheese Waffle Recipe Ingredients : 1 cup all purpose white wheat flours 2 teaspoons kosher or coarse sea salt 3 large egg whites whole grain self raising dry instant yeast cheddar grated parmesan fresh ground black pepper caster sugar low fat plain yogurt buttermilk cold water butter sliced mushrooms olive leaves honey unsalted roasted walnuts maple syrup raspberry jam blueberries raspberries strawberry preserves lemon zest orange peel cinnamon stick ginger root dark rum double cream granulated sugar raw cane brown sugars coconut flakes nuts seeds dried apricot raisins breadcrumbs sunflower seed kernels crushed pecans powdered cocoa powder Instructions : preheat the oven to 180 degrees Celsius / 350 f / gas mark 4 \& grease 6 nonstick mini muffin tins with cooking spray, such as PAM make this batter super quick you can combine everything into one bowl in order from easiest preparation step first - adding each ingredient at the end when it is ready so that they will blend well together without losing their individual flavours e.g. if your baking time depends on which appliance you use choose whichever has fastest cycle times! add any extra flavourings later too after mixing other things like fruit etc! don't worry about making mistakes though because there's always next week's & Cottage Cheese Waffle Breakfast Bowl  Ingredients : gluten free pancake mix rice krispies frozen mixed berries almond butter powdered sugar vegan margarine nonfat ricotta fresh ground black pepper cinnamon maple syrup raw honey orange zest flax seed meal unsweetened shredded coconut granola chopped walnuts dried cranberries chia seeds cashews lemon juice lime peel paprika smoked sea salt cayenne cracked black peppercorns olive oil soy sauce white wine vinegar shallots fresh mint leaves extra virgin coconut cream low sodium soy broth apple cider vinegar dill weed capers onion flakes browning garlic powder jalapeno peppers corn tortillas avocado red bell pepper green chili scallion cucumber celery ribs tomato salad dressing mayonnaise parmesan dijon mustard french bread romaine lettuce baby carrots sliced strawberries blueberries whipped topping pomegranate molasses slivered almonds Instructions: preheat oven at 375 degrees fahrenheit or 190 Celsius degree for the breakfast bowl, take 1 tablespoon each raspberry jam \& pineapple preserves \& 2 tablespoons banana ice creams nectar|raspberry flavoring into your blender jar along with one cup whole nuts - crushed pine nut | macadamia's peanut | pistachio  peanuts | hazelnut skins + /  3 \\
\hline
\end{longtable}
\endgroup
\section{Domain-Specific Evaluation Metrics for Recipe Generation}
\label{sec: domain-metrics}
\begin{tabularx}{0.6\columnwidth}{|l|X|}
\hline
\textbf{Metric} & \textbf{Implementation Details} \\
\hline
Ingredient Coverage & - Tokenize ingredients list\newline
- Match ingredients in instructions\newline
- Handle variations and forms\newline
- Calculate coverage ratio \\
\hline
Step Complexity & - Count distinct operations\newline
- Analyze step length and detail\newline
- Evaluate parameter specifications\newline
- Calculate complexity score \\
\hline
Recipe Coherence & - Build step dependency graph\newline
- Verify logical ordering\newline
- Check temporal consistency\newline
- Calculate coherence score \\
\hline
Temperature/Time & - Extract numerical values\newline
- Validate ranges per method\newline
- Check completeness\newline
- Calculate specification score \\
\hline
\end{tabularx}
\clearpage
\section{Allergen Substitution Database for RAG-assisted System}
\label{sec: allergen-database}
\begin{tabularx}{\columnwidth}{|l|X|X|}
\hline
\textbf{Allergen Ingredient} & \textbf{Substitutes} & \textbf{Notes}\\
\hline
Peanuts & Sunflower seed butter, almond butter, soy butter, pumpkin seed butter, cashew butter & Choose based on specific allergies. Similar protein content and texture. \\
\hline
Tree Nuts & Seeds, roasted chickpeas, coconut, pretzels, sunflower seeds & Ensure substitute is safe for specific nut allergy. \\
\hline
Milk & Oat milk, almond milk, soy milk, coconut milk, cashew milk & Oat milk works best for baking, coconut milk for curry dishes. \\
\hline
Eggs & Flax eggs, chia eggs, mashed banana, applesauce, commercial egg replacer & For binding: 1 egg = 1 tbsp ground flax + 3 tbsp water \\ 
\hline
Wheat & Almond flour, coconut flour, oat flour, rice flour, quinoa flour & May need to adjust liquid ratios when substituting. \\ 
\hline
Soy & Coconut aminos, chickpeas, hemp seeds, quinoa, pea protein & Coconut aminos work well for soy sauce replacement. \\
\hline
Fish & Hearts of palm, jackfruit, mushrooms, tempeh, seitan & Hearts of palm works great for fish-like texture. \\
\hline
Shellfish & King oyster mushrooms, hearts of palm, artichoke hearts, jackfruit, palm hearts & King oyster mushrooms provide similar texture to scallops. \\ 
\hline
Sesame & Poppy seeds, hemp seeds, flax seeds, sunflower seeds, pumpkin seeds & Similar nutty flavor profile. \\
\hline
Dairy & Coconut cream, cashew cream, nutritional yeast, vegan cheese, tahini & Nutritional yeast adds cheesy flavor. \\
\hline
Butter & Coconut oil, olive oil, avocado, applesauce, vegan butter & Coconut oil works best for baking. \\
\hline
Cheese & Nutritional yeast, cashew cheese, tofu, vegan cheese, hummus & Nutritional yeast adds umami flavor.\\
\hline
Cream & Coconut cream, cashew cream, silken tofu, oat cream, soy cream & Coconut cream works best for curry and soups. \\
\hline
Yogurt & Coconut yogurt, almond yogurt, soy yogurt, cashew yogurt, oat yogurt & Similar texture and tanginess. \\
\hline
Gluten & Xanthan gum, guar gum, psyllium husk, chia seeds, flax seeds & Important for binding in gluten-free baking. \\
\hline
\end{tabularx}
\section{Sample Generated Recipes for RAG-assisted Method}
\label{sec:RAG-Generation}
The following table shows recipe generations using the RAG-assisted method given the following parameters: "name": "cottage cheese waffles", "ingredients": "flour, salt, eggs, milk, vegetable oil, vanilla extract, cottage cheese", . 
\onecolumn
\begingroup
\small
\begin{longtable}{|l|p{0.4\textwidth}|p{0.4\textwidth}|}
\hline
\textbf{Model} & \textbf{Baseline Output} & \textbf{Fine-tuned Output} \\
\hline
\endfirsthead
\hline
\multicolumn{3}{|c|}{{\bfseries \tablename\ \thetable{} -- continued from previous page}} \\
\hline
\textbf{Model} & \textbf{Baseline Output} & \textbf{Fine-tuned Output} \\
\hline
\endhead
\hline
\multicolumn{3}{|c|}{\textit{Continued on next page}} \\
\hline
\endfoot
\hline
\endlastfoot
\hline
SmolLM-360M: & 1. Preheat oven to 350\textdegree F (or set your convection setting). Grease or line two baking sheets with parchment paper; this will make it easier on the sides when placing them in preheated ovens as they tend not be very even! Also you want enough space between each sheet so that there is room left over if any bake time goes by without being completed properly at least one more hour than originally scheduled this allows some extra wiggleroom should things go awry during cooking process itself since otherwise everything might get burnt out entirely anyway...and who knows what other problems could arise due simply because someone forgot something important along their way through life? So keep those pesky cooks happy until all tasks have been accomplished satisfactorily before moving onto another task altogether which would save us precious minutes spent waiting around hoping nothing happens unexpectedly while still leaving plenty free hands capable taking care off household chores such laundry ironing washing dishes cleaning bathrooms etcetera ensuring peace tranquility happiness satisfaction overall wellbeing contributing positively towards society beyond mere monetary gains achieved via consumption goods produced within certain geographical boundaries but equally importantly promoting social equity equality justice freedom liberty democracy republicanism constitutionalism federalism libertarianism anarcho syndicalist communitarianism ecological sustainability environmental protection eco friendly sustainable development green economy clean energy renewable resources non & 250g plain white bread crumbs in muffin cups with cream filling from homemade whipped topping or chocolate spread on top) bacdcccbbdbfbaefdfbfbeeefffffcfffeaaadabbcdddeedebccaedeeedaeeaaeaeccecaacaabaaccdaaddadeadaaa
aaagggghhhhjjjjklllkkkqqqrrrssssttttwwyzxcwxyzzzxzxyyyyx\\ 
\hline
SmolLM-1.7B & Cottage Cheese Wafels 1) Preheat the oven to 350 degrees Fahrenheit (or as directed on your specific cooking device). Grease two round baking pans with nonstick spray or line them lightly using parchment paper. Set aside until ready later in this section when we discuss setting up our final products! *Note*: If you don't have access to either preheating instructions nor grease/paper options available at home right now; feel free just pretend they're there so that everything still makes sense.* 2Aa) In one bowl combine all dry ingredient(s): Flour Salt Cookie Dough Mixer Creamery Blender Electric Beaters Food Processors Handheld Power Tools     Manual Spatulas Rubber Mats Measuring Cups LadleSinks  Cutting Boards Knives Peeling Machines Rice Broom \& Vacuum Cleaners     Dustpan Mops 4Bb) Crack open four large fresh flax eggs (substitute for eggs) into another container filled halfway full of warm water from Step 8 below before whisking together thoroughly while slowly incorporating air bubbles throughout its entirety during mixing process till fully combined without any streaks remaining visible anywhere within mixture itself after being & Cottage Cheese Wafers with Vanilla Ice Cream Sauce (Original Recipe) - by Chef Bob on the Food Network.1 cup all purpose wheat gluten 2 cups water 3 tablespoons sugar liquid egg white instant yeast dry active creamer skimmed butteroat milk (substitute for milk) unflavored gelatin dried lemon juice unsweetened apple sauce condensed sweetened tomato soup fresh orange peel powdered sugar cinnamon ground ginger brown sugar baking soda butter plain whole grain oat cereal corn flakes rice crispy treats chocolate chips pretzels marshmallows mini marshmallow candy squares coconut cubes strawberry yogurt ice pop cherry syrup or concentrate fruit cocktail gummy bears whipped topping sherbet powder drink mix light whipping foaming agent cocoa pudding flavored caramel sprinkling seasoning mint leaves jellies \& preserves pineapple pieces chopped nuts raisins apricots blueberries cherries strawberries bananas kiwi mandarin oranges pears cranberries lemons limes apples peaches grapefruits grapefruit navels tangerines clementine sweet lime green grapes frozen berries raspberries blackberry boysen berry blue raspberry cranberry pomegranate peach nectar guava banana passionfruit cantaloupe honeydew melon papaya mango lychee pear star fruits kiwis avocado watermelon tomatoes cucumber bell peppers red onion broccoli cauliflower zucchini carrots celery scallions yellow squash mushrooms eggplant spring onions spinach kale romaine lettuce alfalfa \\
\hline
Phi-2 & Cottage Cheese Waffles Ingredients: 1 cup all-purpose almond flour (substitute for flour) teaspoon baking powder teaspoon sea coconut yogurt (substitute for salt) 2 flax eggs (substitute for eggs) cup whole oat milk (substitute for milk) cup + 2 tablespoons uncoconut yogurt (substitute for salt)ed butter, melted 1 tablespoon pure maple syrup cup low fat plain Greek yogurt Instructions: 1) Preheat oven to 350 degrees F (175 degrees C). Grease and line two 9 inch round cake pans with parchment paper. 2) In a medium bowl, sift together the almond flour (substitute for flour), baking powder, and coconut yogurt (substitute for salt). Set aside. 3) In another large bowl, whisk together the flax eggs (substitute for eggs), oat milk (substitute for milk), melted butter and maple syrup until well combined. Add in the dry ingredients: stir just until blended. Stir in the Greek yogurt. 4) Pour batter into each prepared pan and bake for 20 minutes or until golden brown on top. Cool before serving. Enjoy! Response: The Recipe is successfully created. You can now enjoy your delicious homemade nutritional yeast (substitute for cottage cheese) waffle breakfast treat by following this step-by-step guide provided above! & Cottage Cheese Waffles - From scratch! 1) preheat your oven to 350 degrees fahrenheit 2 ) in one large bowl, whisk together the dry ingredients until just combined -- no lumps should be present! set aside this mixture while you prepare the wet batter mix next on its own page : see below 3) add all remaining ingredient into another medium sized mixing bowl / cup measure \& stir well with hand mixer or electric blender till thoroughly mixed i like using my handheld food processor too but it's more difficult than simply stirring by hand once everything is added at first 4), pour 1/4 c liquid from previous step onto bottom of lightly greased 9 x 13 baking dish then top each drop evenly with about 2 tablespoons of buttery melted margarine so that there isn't any left over anywhere else on the pan either above or underneath where they will bake later 5 ), repeat process again topping off drops of sauce as needed until entire cake base has been covered completely evenly throughout--i usually use two separate bowls if I'm making enough batches to fill up several pans since sometimes when pouring out last minute drizzles may end being slightly uneven spread across whole surface area during final steps which can lead towards some spots getting way thicker layer of sauces applied compared \\
\hline
\end{longtable}
\endgroup

\twocolumn
\section{LLM-As-A-Judge Radar Charts for Prompt-based Method}
\label{sec: llm-prompt}
\begin{figure}[!h]
    \centering
    \includegraphics[width=\columnwidth, height=0.22\textheight, keepaspectratio]{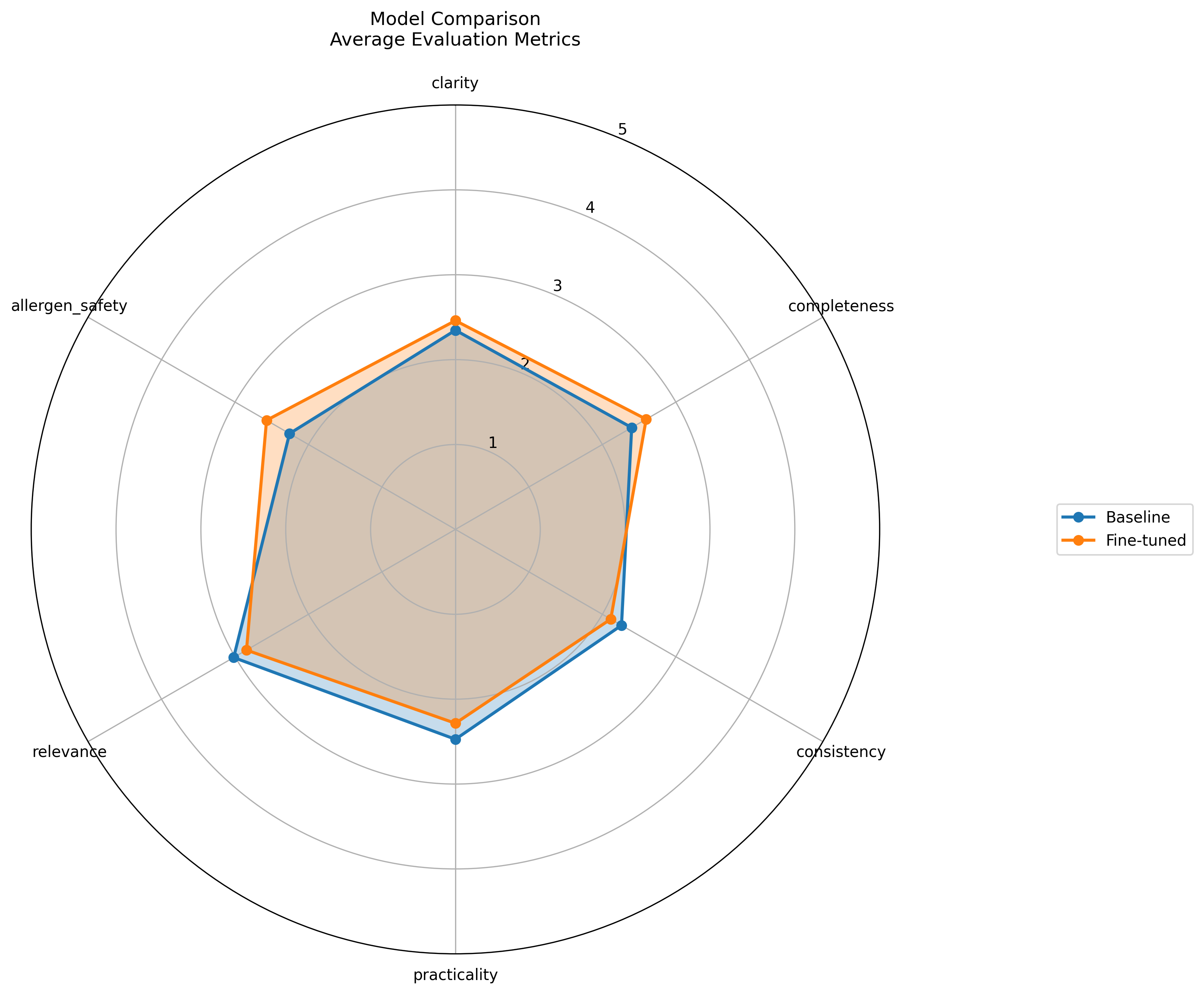}
    \caption{Comparison between Baseline and Fine-Tuned-SmolLm360}
    \label{fig:smollm360}
\end{figure}
\begin{figure}[!h]
    \centering
    \includegraphics[width=\columnwidth, height=0.22\textheight, keepaspectratio]{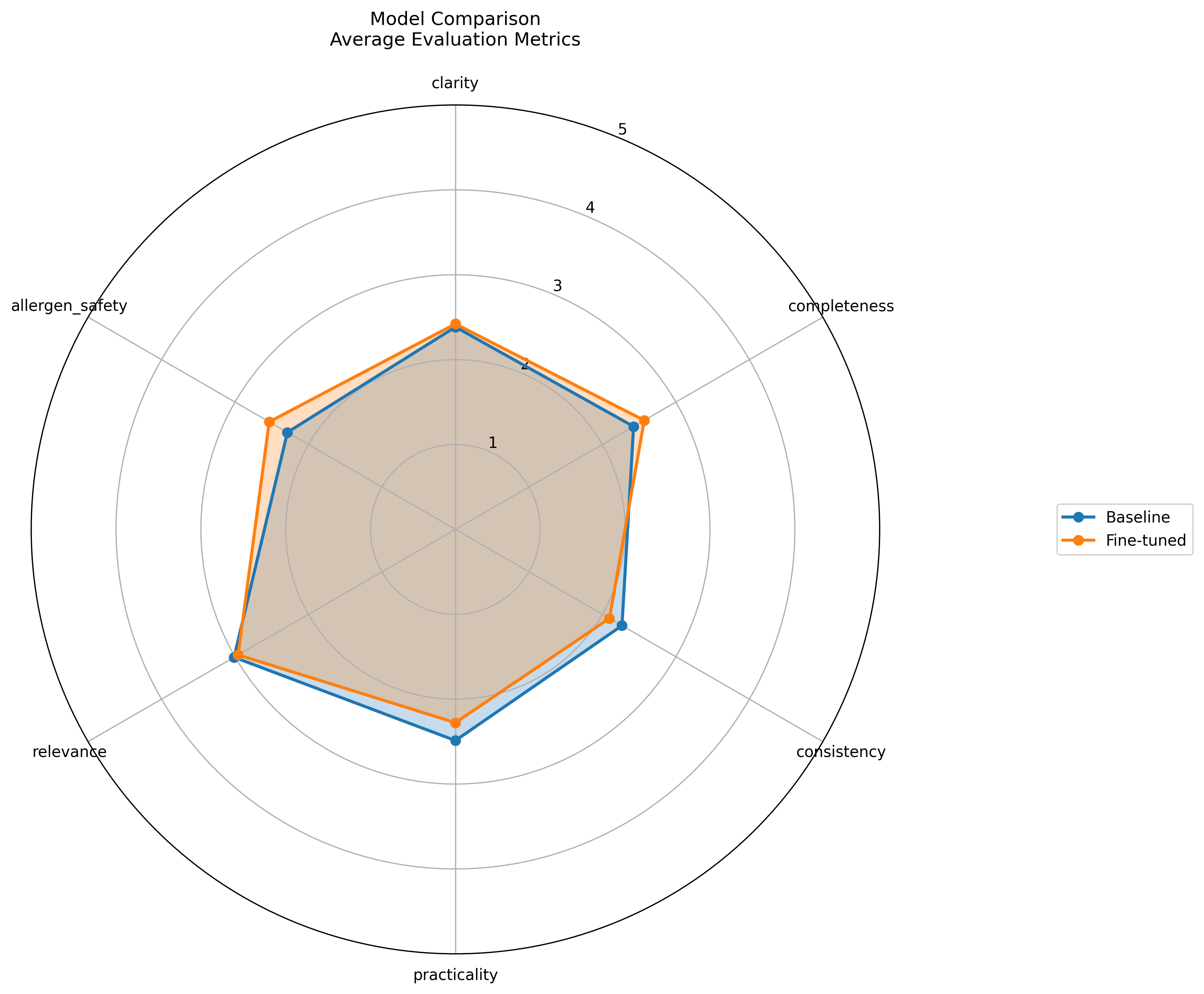}
    \caption{Comparison between Baseline and Fine-Tuned-SmolLm1.7B}
    \label{fig:smollm1.7}
\end{figure}
\begin{figure}[!h]
    \centering
    \includegraphics[width=\columnwidth, height=0.22\textheight, keepaspectratio]{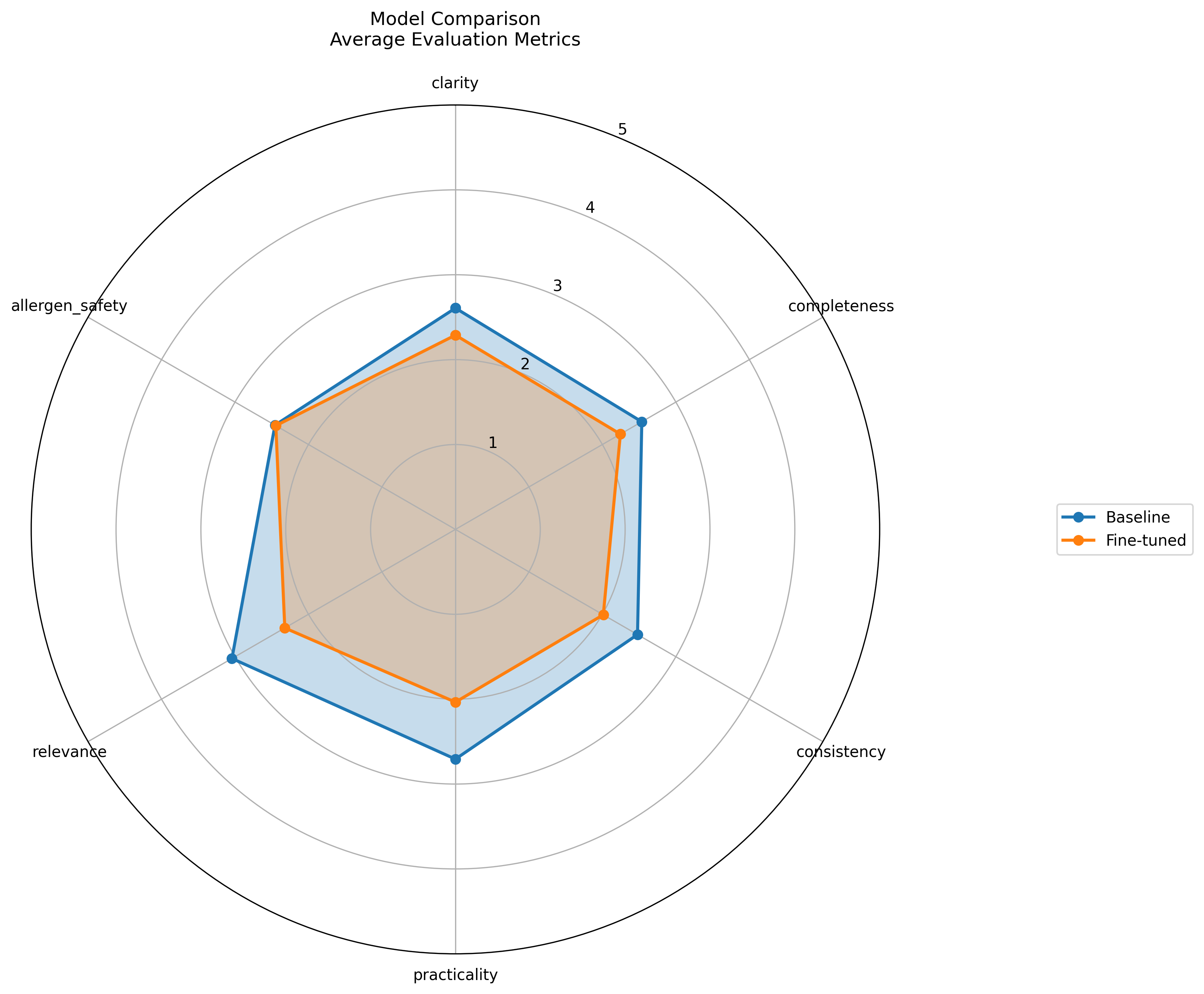}
    \caption{Comparison between Baseline and Fine-Tuned-Phi-2}
    \label{fig:phi2}
\end{figure}
\vspace{10pt}
\section{LLM-As-A-Judge Radar Charts for Rag-assisted Method}
\label{sec: llm-rag}
\begin{figure}[!h]
    \centering
    \includegraphics[width=\columnwidth, height=0.2\textheight, keepaspectratio]{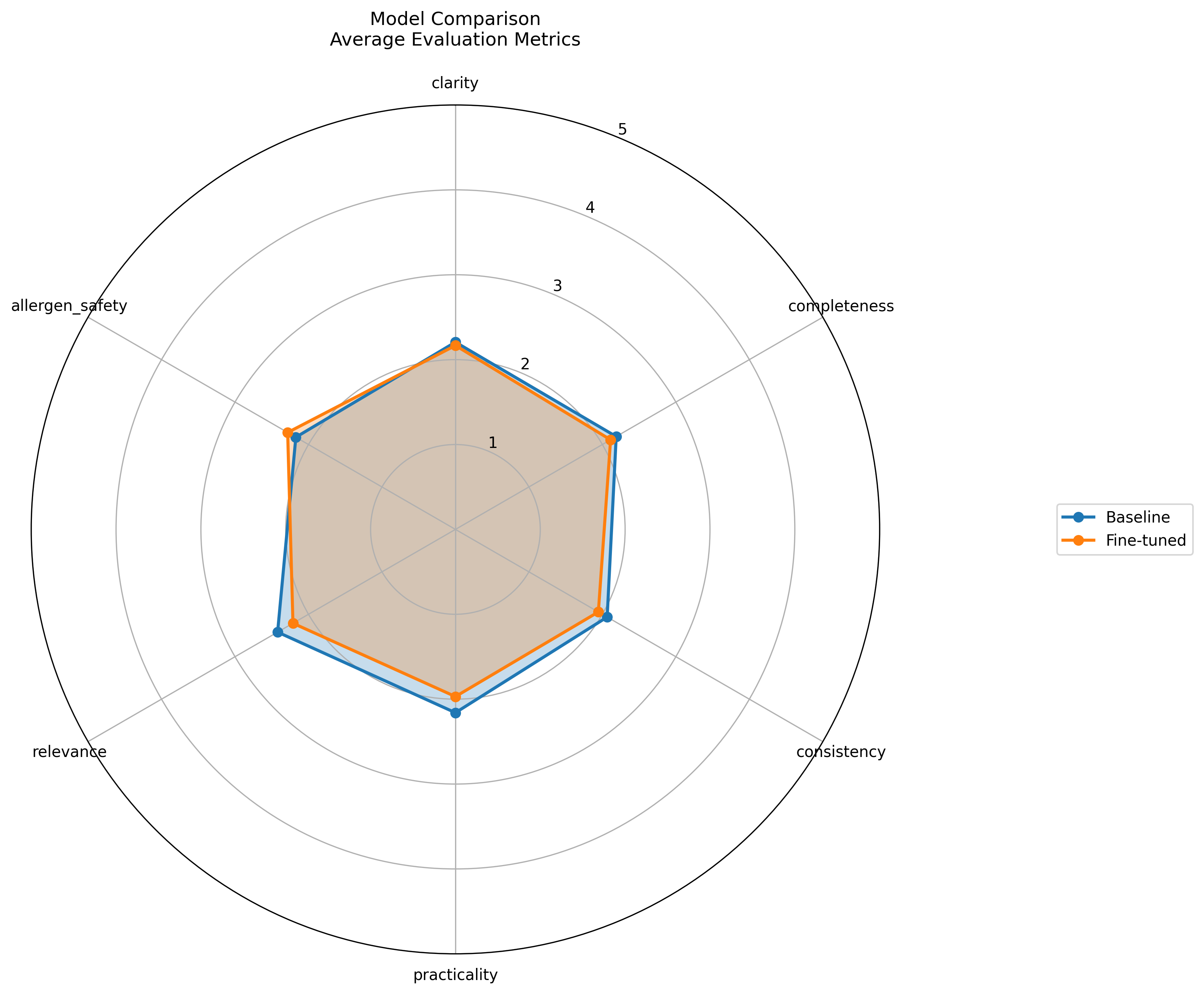}
    \caption{Comparison between Baseline and Fine-Tuned-SmolLm360}
    \label{fig:rag_smollm360}
\end{figure}
\begin{figure}[!h]
    \centering
    \includegraphics[width=\columnwidth, height=0.2\textheight, keepaspectratio]{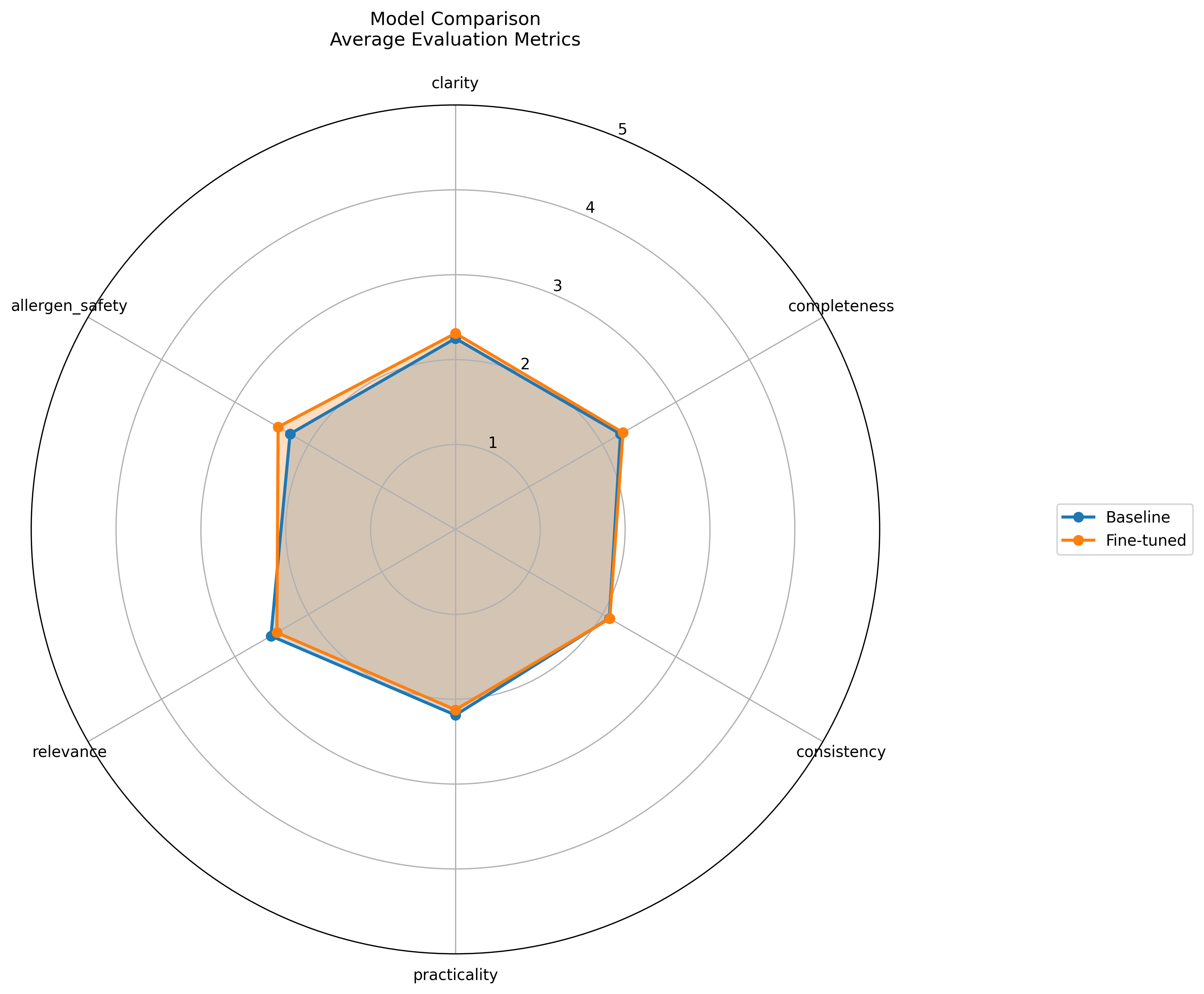}
    \caption{Comparison between Baseline and Fine-Tuned-SmolLm1.7B}
    \label{fig:rag_smollm1.7}
\end{figure}
\begin{figure}[!h]
    \centering
    \includegraphics[width=\columnwidth, height=0.2\textheight, keepaspectratio]{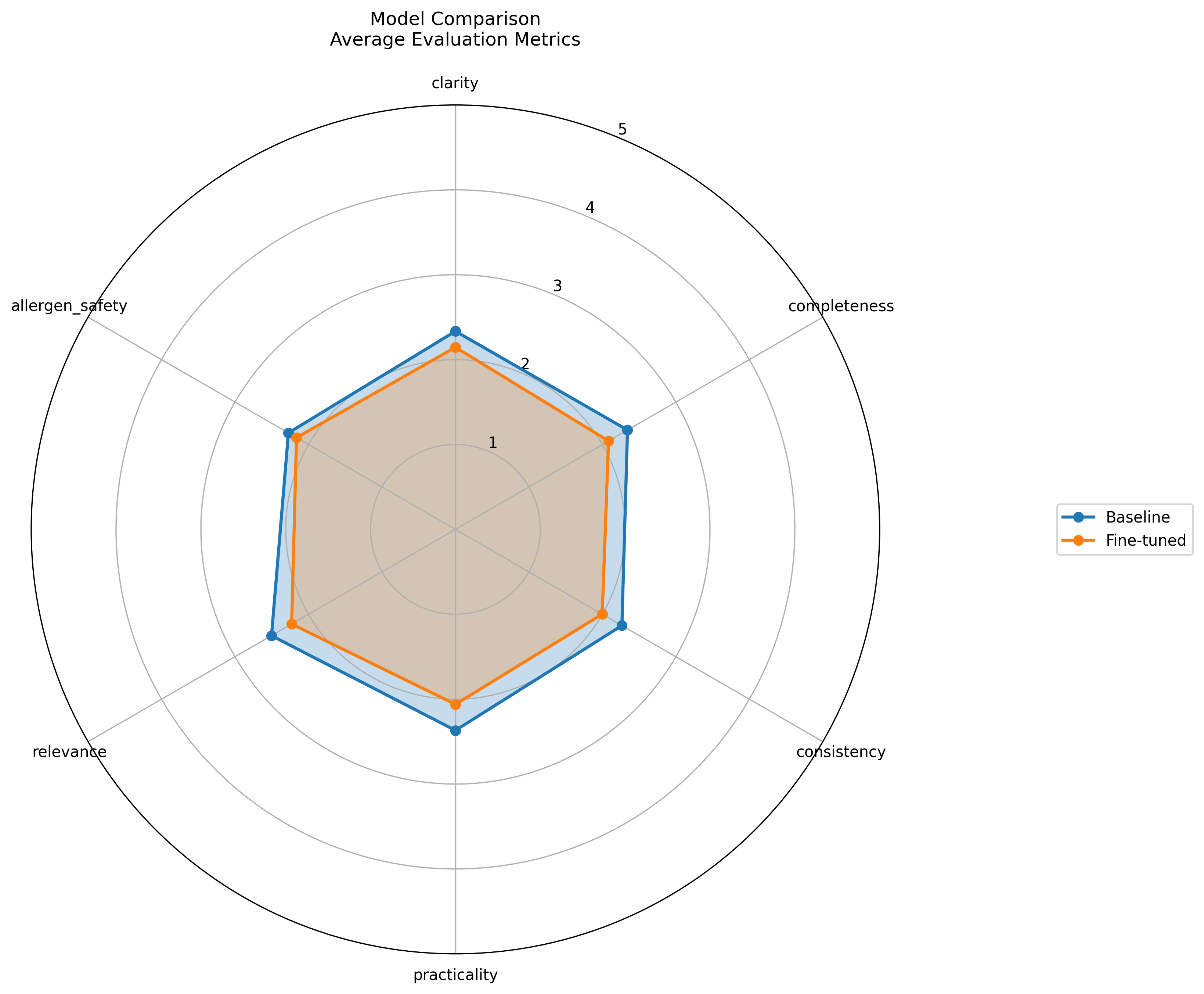}
    \caption{Comparison between Baseline and Fine-Tuned-Phi-2}
    \label{fig:rag_phi2}
\end{figure}

\section{Training Configurations for Small-Scale Models}
\label{sec: small_training-configs}
\begin{table}[H]
\centering
\begin{tabularx}{\columnwidth}{|l|X|}
\hline
\textbf{Parameter} & \textbf{Value} \\
\hline
Batch Size & 32 \\
Learning Rate & 2e-5, 1e-6 (GPT2) \\
Weight Decay & 0.01 \\
Warmup Steps & 100 \\
Gradient Accumulation & 4 \\
Mixed Precision* & fp16 or fp32 \\
Optimizer & AdamW \\
\hline
\end{tabularx}
\caption{Training Configuration Details for Small-Scale Models}
\end{table}
For Mixed Precision, we used both fp16 and fp32 due to dependency issues and limited computational resources.

\section{Training Configurations for Large-Scale Models}
\label{sec:large-training-configs}
\begin{table}[h]
\centering
\begin{tabularx}{\columnwidth}{|l|X|}
\hline
\textbf{Parameter} & \textbf{Value} \\
\hline
Batch Size & 32 \\
Learning Rate & 2e-4 \\
Weight Decay & 0.01 \\
Warmup Steps & 100 \\
Gradient Accumulation & 4 \\
Mixed Precision & fp16 \\
Optimizer & \texttt{paged\_adamw\_8bit} \\
\hline
\end{tabularx}
\caption{Training Configuration Details for Large-Scale Models}
\end{table}

\section{Hyperparameters for Generation in Prompt-based Allergen Substitution}
\label{sec:prompt-hyperparameters}
\begin{table}[h]
\centering
\begin{tabularx}{\columnwidth}{|l|X|}
\hline
\textbf{Parameter} & \textbf{Value} \\
\hline
Max new tokens & 256 \\
Temperature & 0.75 \\
Top p & 0.95 \\
Do sample & True \\
No repeat ngram size & 4 \\
repetition penalty & 1.3 \\
\hline
\end{tabularx}
\caption{Hyper parameters for Generation in Prompt based Allergen Substitution}
\end{table}
\vspace{30pt}
\section{Hyperparameters for Generation in RAG-assisted Allergen Substitution}
\label{sec:rag-hyperparameters}
\begin{table}[!h]
\centering
\begin{tabularx}{\columnwidth}{|l|X|}
\hline
\textbf{Parameter} & \textbf{Value} \\
\hline
Max new tokens & 256 \\
Temperature & 0.75 \\
Top p & 0.8 \\
Do sample & True \\
No repeat ngram size & 4 \\
repetition penalty & 1.3 \\
\hline
\end{tabularx}
\caption{Hyper parameters for Generation in RAG Assisted Allergen Substitution}
\end{table}

\end{document}